\pdfoutput=1

\documentclass[11pt]{article}

\usepackage[final]{acl}

\usepackage{times}
\usepackage{latexsym}
\usepackage{times}  
\usepackage{helvet}  
\usepackage{courier}  

\usepackage{multirow}
\usepackage{adjustbox}
\usepackage{booktabs}
\usepackage{xcolor}
\usepackage{color}
\usepackage{soul} 
\usepackage{caption} 
\usepackage{algorithm}
\usepackage{algpseudocode}
\usepackage{amsmath}
\usepackage{amssymb}

\usepackage{newfloat}
\usepackage{listings}
\usepackage{float}
\usepackage{algpseudocode}

\DeclareCaptionStyle{ruled}{labelfont=normalfont,labelsep=colon,strut=off} 
\floatstyle{ruled}
\newfloat{listing}{tb}{lst}{}

\usepackage[T1]{fontenc}

\usepackage[utf8]{inputenc}

\usepackage{microtype}

\usepackage{inconsolata}

\usepackage{graphicx}

\title{Dynamic Analysis and Adaptive Discriminator for Fake News Detection}

\author{
 \textbf{Xinqi Su\textsuperscript{1,2}},
 \textbf{Zitong Yu\textsuperscript{1}}
 \textbf{Yawen Cui\textsuperscript{3}},
 \textbf{Ajian Liu\textsuperscript{4}},
 \textbf{Xun Lin\textsuperscript{5}},
\\
 \textbf{Yuhao Wang\textsuperscript{1}},
 \textbf{Haochen Liang\textsuperscript{1}},
 \textbf{Wenhui Li\textsuperscript{2}},
 \textbf{Li Shen \textsuperscript{6}},
\\
 \textbf{Xiaochun Cao\textsuperscript{6}}
\\
\\
 \textsuperscript{1}Great Bay University,
 \textsuperscript{2}Tianjin University,
\\
 \textsuperscript{3}The Hong Kong Polytechnic University,
\\
 \textsuperscript{4}Institute of Automation, Chinese Academy of Sciences,\\
 \textsuperscript{5}Beihang University,
 \textsuperscript{6}Sun Yat-sen University
\\
}

\begin{document}
\maketitle
\begin{abstract}
In current web environment, fake news spreads rapidly across online social networks, posing serious threats to society. Existing multimodal fake news detection methods can generally be classified into  knowledge-based and semantic-based approaches. However, these methods are heavily rely on human expertise and feedback, lacking flexibility. To address this challenge, we propose a Dynamic Analysis and Adaptive Discriminator (DAAD) approach for fake news detection. For knowledge-based methods, we introduce the Monte Carlo Tree Search algorithm to leverage the self-reflective capabilities of large language models (LLMs) for prompt optimization, providing richer, domain-specific details and guidance to the LLMs, while enabling more flexible integration of LLM comment on news content. For semantic-based methods, we define four typical deceit patterns: emotional exaggeration, logical inconsistency, image manipulation, and semantic inconsistency, to reveal the mechanisms behind fake news creation. To detect these patterns, we carefully design four discriminators and expand them in depth and breadth, using the soft-routing mechanism to explore optimal detection models. Experimental results on three real-world datasets demonstrate the superiority of our approach. The codes will be released.
\end{abstract}

\section{Introduction}

\begin{figure}[t]
\vspace{-0.8em}
\centering
\includegraphics[width=0.93\columnwidth]{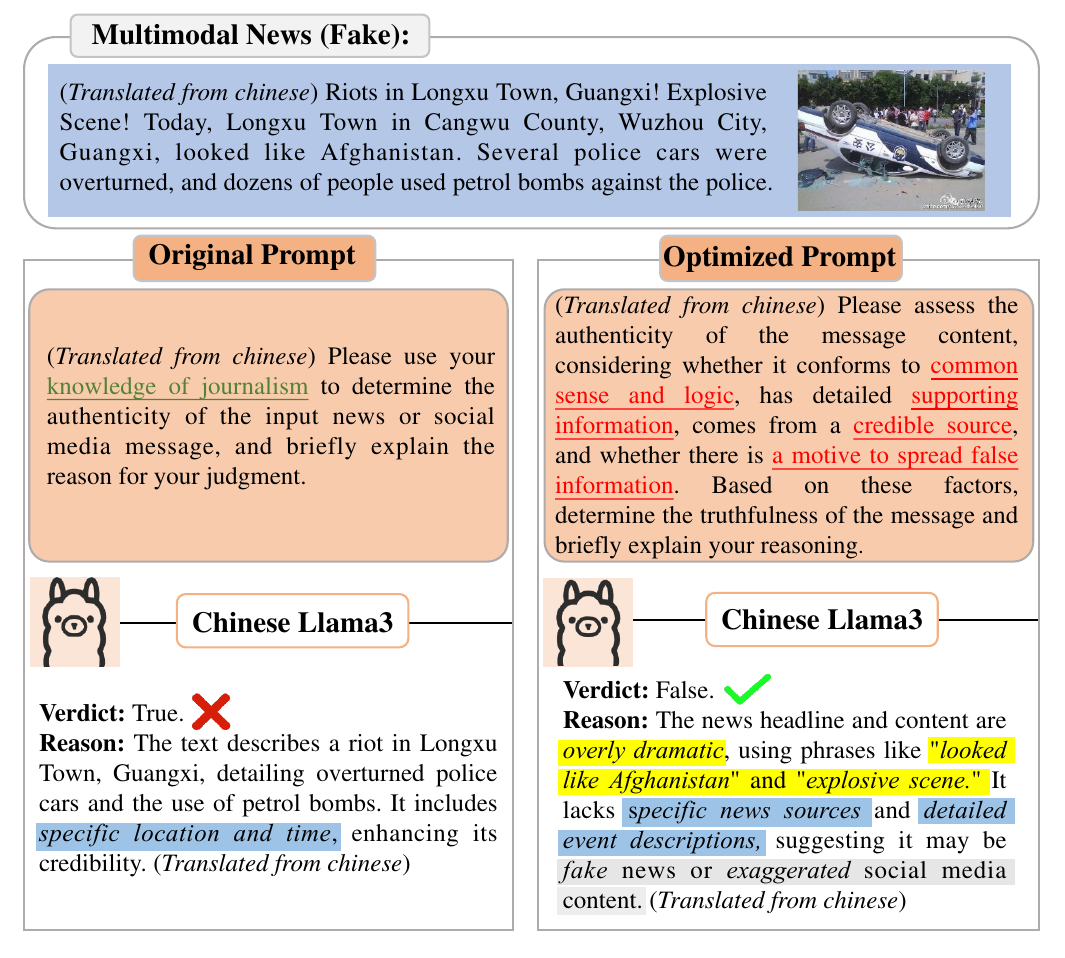} 
\vspace{-1.3em}
\caption{\small{A comparison is made between our optimized prompt vs original prompt using LLMs, such as Chinese LLaMA, for news authenticity classification. The optimized prompt offers richer and domain-specific guidance (highlighted in red), resulting in more accurate comments and predictions (highlighted in gray), such as exaggerations (highlighted in yellow), details (highlighted in blue).}}
\label{fig1}
\vspace{-2.2em}
\end{figure}

News is a crucial channel for the public to understand the world. With the rapid development of online social networks (OSNs) like Twitter and Weibo, the spread of fake news and distorted views has grown exponentially~\cite{willmore2016analysis}. This not only increases the information burden but can also cause panic, leading to significant negative impacts on society~\cite{dong2024unveiling}. To mitigate the adverse effects of fake news, automated fake news detection (FND) has become a research focus, helping network administrators prevent its spread~\cite{ying2023bootstrapping}.

Tweets with images are more engaging, which is why most content on current OSNs is presented in a multimodal format~\cite{wang2023cross,wu2023human}. Additionally, multimodal fake news is more convincing due to its realism, and fake news creators exploit this feature to amplify their influence~\cite{hu2022causal, khattar2019mvae}. Therefore, this paper focuses on multimodal fake news detection (MFND). Many studies have been dedicated to detecting multimodal fake news and mitigating its harmful effects. Current MFND are primarily divided into two categories: (i) semantic-based, and (ii) knowledge-based methods.

\begin{figure}[t]
\centering
\includegraphics[width=0.89\columnwidth]{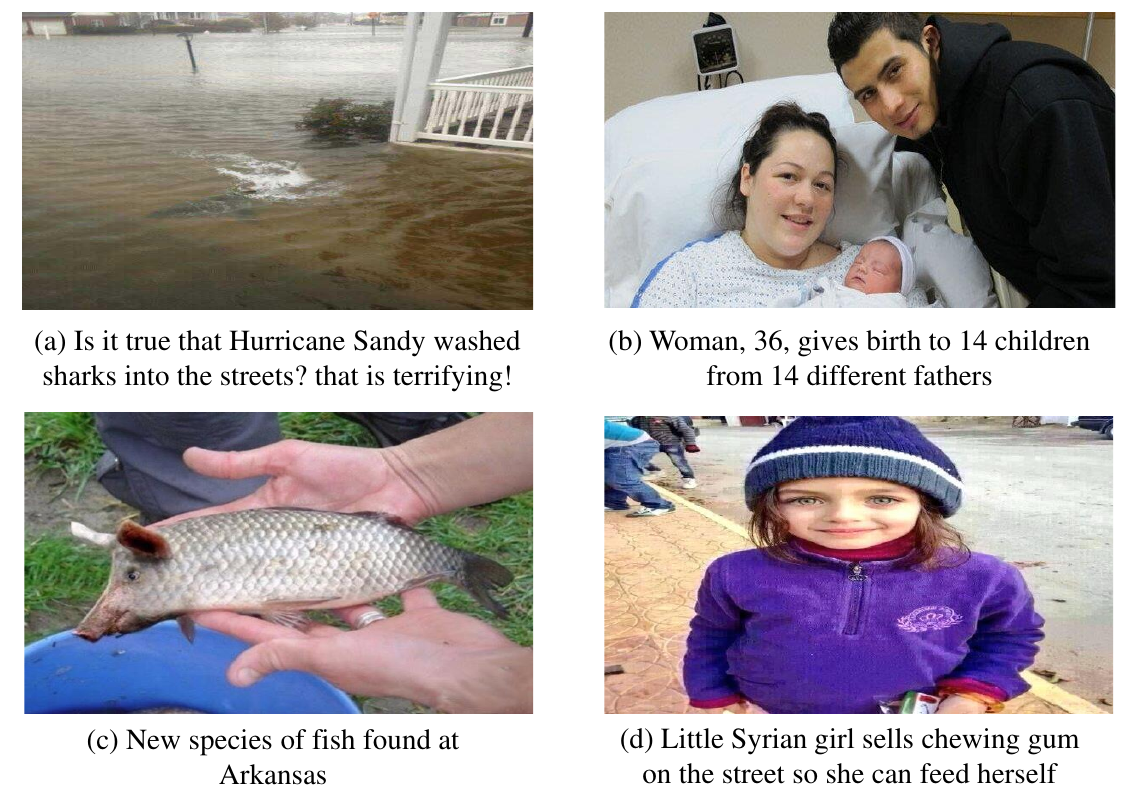} 
\vspace{-0.8em}
\caption{\small{Some examples illustrate the reasons for fake news: (a) the text exhibits clear emotional bias; (b) the text contains exaggerated emotional bias and defies common sense; (c) the image has been manipulated; and (d) there is a semantic inconsistency between the image and the text.}}
\label{fig11}
\vspace{-2.0em}
\end{figure}

 The first category of methods argues that as technology advances, the style of fake news increasingly resembles that of real news, however MFND methods often lack critical information such as social context~\cite{liao2023muser}. To address this, some approaches incorporate knowledge bases (KBs) like Wikipedia~\cite{hu2021compare, tseng2022kahan}, knowledge graphs~\cite{qian2021knowledge,jin2022towards}, and public APIs~\cite{qi2021improving,zhang2021mining} to determine the authenticity of news by capturing consistency with the KB or enriching the news content. However, as fake news becomes more sophisticated, static KBs struggle to provide reliable factual signals. With the emergence of large language models (LLMs) that possess strong reasoning and content understanding capabilities, community feedback indicates that LLMs' comments of news can enhance fake news detection systems~\cite{liu2024fakenewsgpt4,hu2024bad}. Nonetheless, enabling LLMs to perform the intended tasks effectively remains a challenge. For instance, recent studies use prompt engineering to guide LLMs to focus on writing style and common-sense errors in news, allowing them to infer key clues~\cite{liu2024can}. However, there are two major drawbacks: 1) The prompts are manually crafted, overly reliant on human expertise, leading to limited generalization~\cite{ma2024large}; 2) LLMs are highly sensitive to prompt formats, where semantically similar prompts can yield vastly different performances~\cite{kojima2022large, wei2023larger}, and the optimal prompt may be model- and task-specific~\cite{lin2024prompt, hao2024optimizing}. Thus, suitable prompts are crucial for LLMs to effectively analyze news content.

 The second category of methods focuses on capturing the semantic features of news content to distinguish between real and fake news, which can be broadly divided into three subapproaches: 1) Research indicates that there are significant differences in the expression style of credible news versus fake news~\cite{rashkin2017truth,shu2017fake,horne2017just}. Consequently, some methods detect fake news by analyzing elements such as sentiment~\cite{chen2023causal,kwak2020systematic} and logic (e.g., mismatches between semantics and grammar)~\cite{li2024large,xiao2024msynfd}, as illustrated in Fig.~\ref{fig11}a and Fig.~\ref{fig11}b. 2) As shown in Fig.~\ref{fig11}c, some fake news creators use techniques to manipulate images, leading to methods that focus on detecting image manipulation to identify fake news~\cite{lao2024frequency, dong2024unveiling}. 3) As illustrated in Fig.~\ref{fig11}d, another common methods to creating multimodal fake news is the incorrect reuse of outdated images. Therefore, some methods assess the semantic consistency between images and text to detect fake news~\cite{wang2023cross,wu2023see}. Despite the significant performance improvements achieved by these subapproaches from different perspectives~\cite{wu2021multimodal}, two key limitations remain: 1) The detection models are manually crafted and rely on human knowledge and feedback, meaning that the optimal detection models have yet to be fully developed; 2) Existing detection methods are static, applying the same detection model to all samples, which prevents different detection patterns from complementing each other.

\begin{figure*}[h] 
\vspace{-0.8em}
\centering
\includegraphics[width=0.98\textwidth]{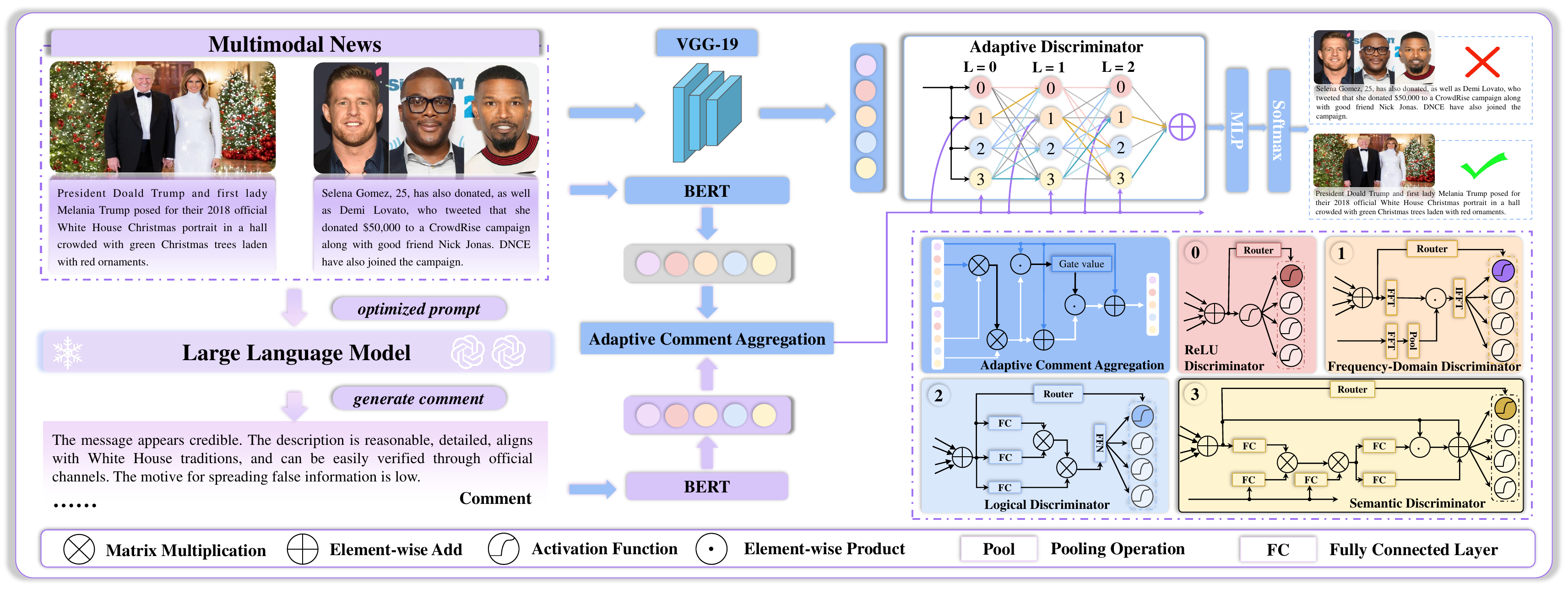} 
\vspace{-1.1em}
\caption{\small{The architecture of our proposed DAAD for fake news detection consists of two main steps: comment on the news by LLMs and adaptive discriminator modeling. First, the news is processed by a prompt-optimized LLM to generate comments, which are then aggregated by the Adaptive Comment Aggregation module. Next, the Adaptive Discriminator module analyzes the cues with various discriminators. Finally, the model outputs the probability that the news is fake.}}
\label{fig3}
\vspace{-1.6em}
\end{figure*}

To address the aforementioned challenges, we propose a Dynamic Analysis and Adaptive Discriminator (DAAD) for fake news detection. First, we introduce the Monte Carlo Tree Search (MCTS) algorithm, strategically leveraging the self-reflective capabilities and error feedback of LLMs to optimize prompts, thereby introducing effective comments. To prevent the optimization process from getting trapped in local minima, we designed a `MemoryBank' component that not only stores historical errors during the search but also compresses them into concise global guidance, encouraging LLMs to adjust prompts from a bird’s eye view. Additionally, we use prompt resampling and batchprompt to expand both the prompt and sample spaces, respectively, further reducing the risk of local minima. Next, we meticulously designed four discriminators to target different deceit patterns: the ReLU and logical discriminator detect the reasonableness of emotion and contextual logic, respectively; the frequency-domain discriminator focuses on identifying image manipulation; and the semantic  discriminator evaluates the alignment between image and text semantics. Finally, we expanded these discriminators in both depth and breadth, combining them with a soft-routing to adaptively explore the optimal detection model.
\vspace{-0.8em}
\begin{itemize}
    \item We propose a joint prompt optimization framework, namely DAAD, based on MCTS and the MemoryBank component to enhance the effectiveness and controllability of LLMs, while mitigating local minima during the optimization process.
    \vspace{-1.0em}
    \item We design four discriminators and employ a soft-routing mechanism to adaptively explore the optimal fake news detection model. 
    \vspace{-1.0em}
    \item Extensive qualitative and quantitative experiments on three benchmarks, \textit{i.e.}, Weibo, Weibo21, and GossipCop, demonstrate the superiority of our approach. 
\end{itemize}


\vspace{-0.8em}
\section{Related Work}
\subsection{Knowledge-based Fake News Detection}
This category of methods employs external knowledge to aid in the detection of fake news. Some approaches use related entities from knowledge graph to enrich news content~\cite{zhang2024reinforced}. Other methods incorporate evidence and capture their consistency to help distinguish between true and fake news~\cite{hu2021compare, qi2021improving}. With the advent of large language models (LLMs), many approaches leverage their powerful knowledge and reasoning capabilities to analyze or enrich news content. For example, FakeNewsGPT4~\cite{liu2024fakenewsgpt4} and LEMMA~\cite{xuan2024lemma} utilize specific knowledge from LLMs to provide supplementary information for fake news detectors. DELL~\cite{wan2024dell} and ARG~\cite{hu2024bad} introduce LLM-based analysis to support decision-making in fake news detection. However, these methods rely on manually crafted prompts, which may result in suboptimal performance. In contrast, our optimized prompts provide LLMs with more detailed and domain-specific guidance, ensuring effective analysis.
\subsection{Content-based Fake News Detection}
This type of approach primarily focuses on the semantic features of news content to detect fake news, including style, image manipulation, and image-text semantic consistency. For instance, MSynFD~\cite{xiao2024msynfd} designed a multi-hop subgraph aggregation mechanism to detect fake news with grammar-semantics mismatches. Some methods utilize sentiment analysis of text or comments to identify fake news~\cite{zhang2021mining, zhang2023sentence, kwak2020systematic}. Other approaches aim to detect image manipulation to recognize fake news~\cite{dong2024unveiling, wu2021multimodal}. Additionally, numerous methods identify fake news by examining the semantic consistency between images and text~\cite{wu2021multimodal, wu2023human, wang2023cross}. In contrast, our method explicitly extends different detection models in both depth and breadth, using dynamic routing mechanism to adaptively guide them for fake news detection.

\section{Methodology}
The architecture of our proposed DAAD, as shown in Fig.~\ref{fig3}, predicts the label based on the multimodal news content and the comment provided by the LLM.\\
\textbf{\textit{Definition 1}} (Multimodal Fake News Detection): \textit{A piece of multimodal news is defined as $ \mathcal{N} = \{\mathcal{I}, \mathcal{T}\} $, where $\mathcal{I}$ and $\mathcal{T}$ represent the image and text of the news, respectively. Our goal is to learn a probability distribution $ P(\mathcal{Y}|\mathcal{I}, \mathcal{C}) $ that can effectively distinguish between fake and real news, where $ \mathcal{Y} = 1 $ indicates fake news, $ \mathcal{Y} = 0 $ indicates real news, and $\mathcal{C}$ denotes the comments from LLMs about the news. }
\subsection{Learning Optimized Prompts}
Here, we will detail the process of prompt optimization to generate expert-level prompts in the news domain and produce insightful comments.\\
\textbf{\textit{Definition 2}} (Prompt Optimization): \textit{Given a LLM $ \mathcal{M} $, our goal is to find a prompt $\rho$ that generates the best comments $ \mathcal{C}_\diamondsuit$ for a small set of sampled news $ \mathcal{N}_\diamondsuit = \{\mathcal{T}_\diamondsuit \} $. Formally, this can be framed as an optimization problem. We aim to identify a prompt $ \rho $ within the natural language space $\mathcal{S}$ when $ \mathcal{M} $ is prompted with $\rho$ and $\mathcal{T}_\diamondsuit $, the expected value of each sample's comment score $ \phi (\rho, \mathcal{T}_\diamondsuit) $ is maximized.}
\begin{equation} \label{eqn1}\small
\rho^\star = \arg\max_{\rho \in \mathcal{S}} \, \mathbb{E}_{\mathcal{T}_\diamondsuit} [\phi(\rho, \mathcal{T}_\diamondsuit)].
\end{equation}
\subsubsection{The Markov Decision Process (MDP)}
We employ the reinforcement learning (RL)~\cite{li2017deep} for prompt optimization, modeling it as a MDP represented by a 4-tuple $(\mathcal{S}, \mathcal{A}, \Gamma, \Upsilon)$, where $\mathcal{S}$ denotes the state space, $\mathcal{A}$ represents the action space, $\Gamma$ is the state transition function $\Gamma: \mathcal{S} \times \mathcal{A} \to \mathcal{S}$, and $\Upsilon$ is the reward function $\Upsilon: \mathcal{S} \times \mathcal{A} \to \mathcal{R}$. To incorporate domain-specific knowledge into prompt revisions meaningfully, we propose action generation based on error feedback. Fig. \ref{figllm}a illustrates a state transition process. In the current state $s_t$, the agent generates an action $a_t$ based on $a_t \sim p_{\mathcal{M}'}(a \mid s_t, \mathcal{M'})$, where $\mathcal{M'}$ is a LLM used for action generation. Subsequently, the agent obtains the new state $s_{t+1}$ based on the transition function $s_{t+1} \sim p_{\mathcal{M}''}(s \mid s_t, a_t, \mathcal{M''})$, where $\mathcal{M''}$ is a LLM used for state transition. The reward $\Upsilon$ is defined as the accuracy on the validation set $\mathcal{N}_\diamondsuit$.

\subsubsection{Enhancing Prompt Optimization}
To enhance the agent's exploration of the prompt $\rho$, we employ Monte Carlo Tree Search (MCTS) for optimization, inspired by~\cite{wang2023promptagent}. Specifically, MCTS maintains a state-action value function $Q: \mathcal{S} \times \mathcal{A} \to \mathcal{R}$ to estimate the potential rewards for state-action pairs along various paths. It iteratively performs selection, expansion, simulation, and backpropagation to update $Q$ and expand the tree. After a predefined number of iterations, the path with the highest reward is selected as the final prompt $\rho$.

\textbf{Selection:} At each iteration, the process starts from the root node $s_0$ and traverses each layer of the tree, using Upper Confidence bounds applied to Trees (UCT) to select child nodes until reaching a leaf node.
\begin{equation}\label{eqn2}\small
a_t^*=\underset{a_t^{\prime} \in \mathcal{A}\left(s_t\right)}{\arg \max }\left(Q\left(s_t, a_t^{\prime}\right)+k \cdot \sqrt{\frac{\ln N \left(s_t\right)}{N\left(\operatorname{ch}\left(s_t, a_t^{\prime}\right)\right)}}\right) ,
\end{equation}
where $ A(s_t) $ and $ N(s_t)$ represent the action set and visit count of node $ s_t$,  respectively, while $ ch(s_t, a'_t) $ denotes the child node generated by action $ a'_t $. The constant $ k $ is a hyperparameter. $Q(s_t, a_t^{\prime})$ denotes the average accuracy/reward of node $s_t$ on the validation set $\mathcal{N}_\diamondsuit$.




\begin{figure}[t]
\centering
\includegraphics[width=0.98\columnwidth]{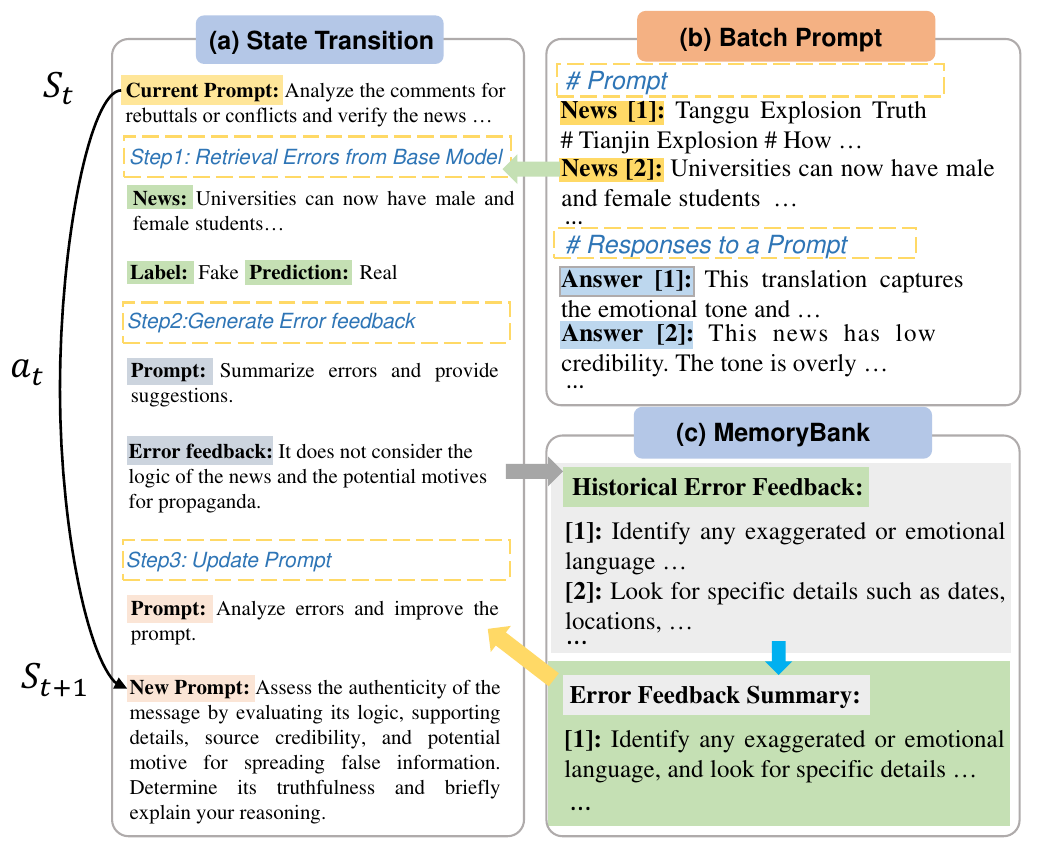} 
\vspace{-1.3em}
\caption{\small{(a) An example of a state transition. Given a prompt ($s_t$), Step 1 is responsible for collecting errors, Step 2 provides modification suggestions ($a_t$) through a LLM, and Step 3 generates the new prompt ($s_{t+1}$) through other LLM. (b) An example of Batch Prompt. (3) The MemoryBank collects historical errors and summarizes them into global guidance, which is fed back into the state transition process.}}
\label{figllm}
\vspace{-1.5em}
\end{figure}

\textbf{Expansion:} Through the MDP, new child nodes are added to the selected leaf node. However, during expansion, the number of samples used to generate error feedback (action) are limited, making it particularly prone to injecting instance-specific details into the prompt (state).

To address this limitation, we developed \textit{Batchprompt}, as shown in Fig. \ref{figllm}b, which processes multiple samples in a single prompt to generate multiple error feedbacks. Additionally, as illustrated in Fig. \ref{figllm}c, we created a \textit{MemoryBank} that not only stores a large volume of historical error feedback but also uses LLM to summarize errors with the prompt, “Summarize the reasons and suggestions, focusing on key information.” This forms a hierarchical memory, providing global guidance. Third, to increase the probability of generating successful prompts, we not only consider the current best prompt but also perform local exploration around it through \textit{Resampling}. Specifically, we instruct the LLM to generate semantically similar variants with the prompt, “Generate ${num}$ variants of the prompt while preserving the original meaning.”

\textbf{Simulation:} It is used to simulates the future trajectory of the selected node during expansion. To simplify, we directly generate multiple actions and select the node with the highest reward to proceed to the next layer of the tree structure.

\textbf{Backpropagation:} Upon reaching the maximum depth, we backpropagate the rewards along the path from the root node to the terminal node and update the $Q$  function.
\begin{equation}\small
Q^*\left(s_t, a_t\right)=\frac{1}{Z} \sum_{\zeta=1}^Z\left(\sum_{s^{\prime} \in S_{s_t}^\zeta, a^{\prime} \in A_{a_t}^\zeta} \Upsilon\left(s^{\prime}, a^{\prime}\right)\right) ,
\label{eq3}
\end{equation}
    where $Z$ represents the number of future trajectories originating from $s_t$, $S_{s_t}^\zeta$ and  $A_{a_t}^\zeta$ denote the $\zeta$-th state and action list in these trajectories, respectively. The MCTS algorithm and the specific meta-prompts for LLM, MemoryBank, BatchBrompt, and Resampling can be found in Appendix ~\ref{po}.
\subsection{Adaptive Discriminator Modeling}
\subsubsection{Feature Representation}
Given an image $\mathcal{I}$, we extract visual features using VGG-19~\cite{simonyan2014very}. These features are then mapped to a $d$-dimensional space via a fully connected (FC) layer. The features for visual are represented as $V = \left[v_1, v_2, \ldots, v_r\right] \in \mathbb{R}^{r \times d} $. For a given text $\mathcal{T}$ and comment $\mathcal{C}$, we use a pre-trained BERT~\cite{devlin2018bert} to extract word embeddings. These embeddings are also mapped to a $d$-dimensional space through a FC layer. The features for text and comment are represented as $T = \left[t_1, t_2, \ldots, t_n\right] \in \mathbb{R}^{n \times d}$ and $C = \left[c_1, c_2, \ldots, c_m\right] \in \mathbb{R}^{m \times d} $, respectively.

\subsubsection{Adaptive Comment Aggregation Module}
Due to issues such as hallucination resulting negative comments from LLM, which can affect training and inference, we propose an Adaptive Comment Aggregation Module, it includes two components: Comment Aggregation and Gated Fusion.

\textbf{Comment Aggregation:} We use text to identify key information in comments, facilitating positive passing between text and comments. Firstly, compute the affinity matrix between them.
\begin{equation}\label{4}\small
{\bf{A}} = {\left( {{{{\bf{ W}}}_{\rm{c}}}C} \right) }\left( {{{{\bf{ W}}}_t}{T}} \right)^ \top,
\end{equation}
where \( W_c \) and \( W_t \) are learnable weight matrices. The aggregated information flow from the comments is summarized using the affinity matrix \( \bf{A} \):
\begin{equation}\label{5}\small
\tilde{C} = \text{softmax}\left(\frac{{\bf{A}}^\top}{\sqrt{d}}\right) C .
\end{equation}

Next, we perform comment fusion.

\textbf{Gated Fusion:} We calculate the corresponding gate as:
\begin{equation}\label{6}\small
{g_i} = \sigma ({t_i} \odot \tilde c_i ),i \in \{ 1,...,n\} ,
\end{equation}
where \(\odot\) denotes the element-wise product, \(\sigma(\cdot)\) represents the activation function, and \(g_i \in \mathbb{R}^d\) is the fusion gate, which promotes the integration of positive comments while suppressing negative ones. All gates can be represented as \(G_t = [g_1, \cdots, g_n] \in \mathbb{R}^{n \times d}\). Meanwhile, to retain the original feature of text that not requiring fusion, the fused features are further combined with the original features through residual connections:
\begin{equation}\label{7}\small
\hat T = {W_f}({G_t} \odot (T \oplus {{\tilde C} })) + T ,
\end{equation}
where \(W_f\) is a learnable weight matrices, and \(\oplus\) denotes element-wise add.

\subsubsection{Discriminator}
As mentioned above, we design four discriminators to detect different deception patterns. Since both images and text exhibit these patterns, we implement image-based version here, while the text-based version and the motivation behind the discriminator can be found in the Appendix~\ref{text-based}.

\textbf{ReLU Discriminator:} Intuitively, complex discriminators are unnecessary for detecting fake news that includes exaggerated emotional words and punctuation, and other obvious cues. Instead, we designed a simple ReLU-based discriminator to retain essential distinguishing features, formulated as: $\mathcal{H}_0(V)=\operatorname{ReLU}(V)$.

\textbf{Frequency Domain Discriminator:} To detect manipulated images or texts, we apply the Fast Fourier Transform (FFT) to convert spatial domain features into the frequency domain, enabling the detection of anomalous frequency components and identifying potential manipulations.
\begin{equation}\label{8}\small
\mathcal{V}[k] / \mathcal{T}[k] = \mathcal{F}_{\text{fft}}\left( v_i / \hat{t}_i \right) = \sum_{i=0}^{n-1/m-1} \left( v_i / \hat{t}_i \right) e^{-j 2\pi \frac{ki}{n/m}} ,
\end{equation}
where $\mathcal{V} \in \mathbb{R}^{r \times d}$ and $\mathcal{T} \in \mathbb{R}^{n \times d}$ represent the complex vectors, and $\mathcal{V}_{[k]}$ and $\mathcal{T}_{[k]}$ are the spectrum of $v_i$ and $\hat{t}_i$ at the frequency $\frac{2\pi k}{m}$ and $\frac{2\pi k}{n}$, respectively. Inspired by~\cite{lao2024frequency}, we enhance the image modality discrimination cues through textual features while suppressing irrelevant information. First, the frequency features $\mathcal{T}$ are pooled, and then a Temporal Convolutional Network (TCN) is used to extract important information from the text context, thereby filtering out irrelevant discriminative cues in the images:
\begin{equation}\label{9}\small
\mathcal{\hat{V}} = \mathcal{V} \odot TCN\left( Avg\left( \mathcal{T} \odot W_x \right) \right) ,
\end{equation}
where $W_x$ is a trainable parameter matrix. Finally, we use the Inverse Fast Fourier Transform (IFFT) to restore the frequency domain features of image back to the spatial domain:
\begin{equation}\label{10}\small
\tilde{V} \leftarrow \mathcal{F}_{\text{fft}}^{-1}(\mathcal{\hat{V}}) .
\end{equation}

Therefore, the above operations can be summarized as: $\mathcal{H}_1(V, \hat{T}) = \left[ \tilde{v}_1, \ldots, \tilde{v}_r \right]$.

\textbf{Logical Discriminator}: It is used to assess the contextual consistency between images and text (e.g., semantic and syntactic coherence of the text or global consistency of image features). We employ the multi-head attention mechanism to capture contextual dependencies across different subspaces feature of the image.

\begin{equation}\label{10}\small
\operatorname{MultiHead}(V)=\operatorname{Concat}\left(\operatorname{head}_1, \ldots, \text { head }_h\right) W_o ,
\end{equation}
where the  $h$-th head is represented as  $\text{head}_h$, and $ W_O$ denotes a learnable parameter matrix. Each attention head performs self-attention by computing the query $(Q = w_q \times V )$, key $(K = w_k \times V )$, and value $(V = w_v \times V )$ vectors:
\begin{equation}\label{11}\small
Attention(Q, K, V) = softmax\left(\frac{QK^T}{\sqrt{d}}\right)V .
\end{equation}

Next, we use a feed-forward network (FFN) to combine the features produced by different head. The above operations can be summarized as: $\mathcal{H}_2(V)=\operatorname{FFN}(\text {MultiHead}(V))$.

\begin{table*}[t]
\centering
\caption{\small{The results of the comparison between the proposed DAAD method and mainstream approaches on the Weibo, Weibo-21, and GossipCop datasets. The best performances are highlighted in bold.}}
\vspace{-0.8em}
\begin{adjustbox}{width=\textwidth}
\begin{tabular}{lcccccccc}
\toprule
\multirow{2}{*}{Dataset} & \multirow{2}{*}{Method} & \multirow{2}{*}{Accuracy} & \multicolumn{3}{c}{Fake News} & \multicolumn{3}{c}{Real News} \\ \cline{4-9}
                         &                         &                           & Precision & Recall & F1-score & Precision & Recall & F1-score \\ \hline
\multirow{8}{*}{Weibo}   & EANN(SIGKDD 2018)\cite{wang2018eann}          & 0.827 & 0.847 & 0.812 & 0.829 & 0.807 & 0.843 & 0.825 \\
                         & MCAN(ACL Findings 2021)\cite{wu2021multimodal}      & 0.899 & 0.913 & 0.889 & 0.901 & 0.884 & 0.909 & 0.897 \\
                         & CAFE(WWW 2022)\cite{chen2022cross}         & 0.840 & 0.855 & 0.830 & 0.842 & 0.825 & 0.851 & 0.837 \\
                         & NSLM(AAAI 2024)\cite{dong2024unveiling}      & 0.885 & -     & -     & -     & -     & -     & -     \\
                         & RaCMC(AAAI 2025)\cite{yu2024racmc} & 0.915 & 0.910 &\textbf{0.924} & 0.917 & 0.921 & 0.906 & 0.914 \\
                         & \textbf{DAAD (Ours)}                                  & \textbf{0.932} & \textbf{0.942} & 0.915 & \textbf{0.928} & \textbf{0.922} & \textbf{0.947} & \textbf{0.934} \\ \hline
\multirow{5}{*}{Weibo-21} & EANN(SIGKDD 2018)\cite{wang2018eann}          & 0.870 & 0.902 & 0.825 & 0.862 & 0.841 & 0.912 & 0.875 \\
                         & SpotFake(AAAI 2020)\cite{singhal2020spotfake+} & 0.851 & \textbf{0.953} & 0.733 & 0.828 & 0.786 & 0.964 & 0.866 \\
                         & CAFE(WWW 2022)\cite{chen2022cross}         & 0.882 & 0.857 & 0.915 & 0.885 & 0.907 & 0.844 & 0.876 \\
                         & BMR(AAAI 2023)\cite{ying2023bootstrapping}  & 0.929 & 0.908 & \textbf{0.947} & 0.927 & \textbf{0.946} & 0.906 & 0.925 \\
                         & \textbf{DAAD (Ours)}                                  & \textbf{0.942} & 0.951 & 0.925 & \textbf{0.938} & 0.931 & \textbf{0.955} & \textbf{0.943} \\ \hline
\multirow{7}{*}{GossipCop} & EANN(SIGKDD 2021)\cite{wang2018eann}         & 0.864 & 0.702 & 0.518 & 0.594 & 0.887 & 0.956 & 0.920 \\
                         & CAFE(WWW 2022)\cite{chen2022cross}         & 0.867 & 0.732 & 0.490 & 0.587 & 0.887 & 0.957 & 0.921 \\
                         & BMR(AAAI 2023)\cite{ying2023bootstrapping} & 0.895 & 0.752 & 0.639 & 0.691 & 0.920 & \textbf{0.965} & 0.936 \\
                           & AKA-Fake(AAAI 2024)\cite{zhang2024reinforced} & 0.856 & -     & -     & -     & -     & -     & -     \\
                         & RaCMC(AAAI 2025)\cite{yu2024racmc} & 0.879 & 0.745 & 0.563 & 0.641 & 0.902 & 0.954 & 0.927 \\
                         & \textbf{DAAD (Ours)}                                  & \textbf{0.904} & \textbf{0.760} & \textbf{0.708} & \textbf{0.734} & \textbf{0.932} & 0.947 & \textbf{0.940} \\ \bottomrule
\end{tabular}
\end{adjustbox}
\label{tab1}
\vspace{-1.3em}
\end{table*}

\textbf{Semantic Discriminator:}
It discriminates fake news by identifying fine-grained differences between image and text using cross-attention. First, compute the contextual vectors of the text that correspond to the image segments:
\begin{equation}\label{13}\small
{T^v} = softmax\left( {\frac{{{{\left[ {{W_q}V} \right]} }\left[ {{W_k}\hat{T}} \right]^ \top}}{{\sqrt d }}} \right)\left[ {{W_v}\hat{T}} \right] .
\end{equation}

To finely identify differences between image and text, we refine the discriminative cues between image and text features. First, map the contextual vector ${T^v}$ to generate scaling vector $\beta_v$ and shifting vector $\gamma_v$.
\begin{equation}\small
\left\{ {\begin{array}{*{20}{l}}
{{\gamma _v} = \sigma \left( {{W_\gamma }\left( {{T^v}} \right)} \right),}\\
{{\beta _v} = {W_\beta }\left( {{T^v}} \right),}
\end{array}} \right.
\label{eq14}
\end{equation}
where $W_\gamma$ and $W_\beta$ is learnable parameter matrices and $\sigma$ is activation function. Then, we compute the refined features using $\gamma_v$, $\beta_v$, MLP and residual connections.
\begin{equation}\small
\bar{V}_r=M L P\left(V \odot \gamma_v +\beta_v\right)+V .
\end{equation}

The above operations can be summarized as: $\mathcal{H}_3(V,\hat{T})=  \left[ \bar{v}_1, \ldots, \bar{v}_r \right]$.
\subsubsection{Soft Router}
To adaptively utilize different discriminators for exploring optimal detection models, we expanded the proposed discriminators both in depth and width. The input image feature for the $i$-th discriminator is:
\begin{equation}\small
V_i^{(l)} = \begin{cases} 
V, & l = 0, \\
\sum_{j=0}^{\mathcal{O}-1} \tau_{j, i}^{(l-1)} \mathcal{H}_j^{(l-1)}, & l > 0, 
\end{cases}
\end{equation}
where \(\mathcal{O}\) represents the total number of discriminators in each layer, \(\mathcal{H}_j^{(l-1)} \in \mathbb{R}^{r \times d}\) denotes the output of the \(j\)-th discriminator in the \((l-1)\)-th layer, and \(\tau_{j, i}^{(l-1)}\) indicates the path probability from the \(j\)-th discriminator in the \((l-1)\)-th layer to the \(i\)-th discriminator in the \(l\)-th layer:
\begin{equation}\small
\tau_i^{(l)}=\operatorname{ReLU}\left\{ \sigma \left[M L P\left(Avg({V_i}^{(l)})\right)\right]\right\},
\end{equation}
when routing ends, the output feature is $\mathcal{X} = V^l$. To preserve the discriminative cues, we apply softmax pooling to $\mathcal{X}$ to obtain the final feature $\mathcal{X^*}$, which is then fed into the prediction layer to output the probability of the news being fake~\cite{wu2021multimodal}.
\begin{equation}\small
\hat{y}=\operatorname{softmax}\left(\max \left(0, \mathcal{X^*} W_s'\right) W_s\right),
\end{equation}
where $W_s'$ and $W_s$ are learnable parameters.
\subsection{Objective Function}
For balanced datasets like Weibo and Weibo21, we use standard cross-entropy loss for training. For the imbalanced dataset like GossipCop, we have improved the cross-entropy loss accordingly.
\begin{equation}\small
\mathcal{L}=-\frac{1}{N} \sum_{i=1}^Ny_i \log \left(\hat{y} \right)+\left(1-y_i\right) \hat{y}^\lambda \log \left(1-\hat{y}\right)
\label{cross}
\end{equation}

In $ (1 - y_i) \hat{y}^\lambda \log (1 - \hat{y}) $, the $\hat{y}^ \lambda$ reduces the loss value for easily recognizable true news, thus compelling the training process to focus more on challenging true and false news samples.

\vspace{-0.6em}

\section{Experiments}
\subsection{Experiment Setting} Experiments were conducted on three widely used datasets: Weibo~\cite{jin2017multimodal}, Weibo-21~\cite{nan2021mdfend}, and GossipCop~\cite{shu2020fakenewsnet}. The Weibo dataset consists of 3,783 fake news and 3,749 real news for training, and 1,000 fake news and 996 real news for testing. The Weibo-21 dataset contains 4,487 fake news and 4,640 real news. For fair comparison, the dataset was split into training and testing sets at a 9:1 ratio, following the previously established protocol~\cite{ying2023bootstrapping}. The GossipCop dataset is imbalanced, with 2,036 fake news and 7,974 real news for training, and 545 fake news and 2,285 real news for testing. For prompt optimization, 30\% of the samples from each dataset were randomly selected for validation. More details of the implementation can be found in the appendix ~\ref{imple}.

\subsection{Performance Comparison}
Table~\ref{tab1} shows a comparison of our DAAD with mainstream methods. The accuracy of DAAD on Weibo, Weibo-21, and GossipCop is 93.2\%, 94.2\%, and 90.4\%, respectively, representing improvements of 1.7\%, 1.3\%, and 0.9\% over state-of-the-art models. Specifically, on the Weibo dataset, DAAD achieves the best performance across nearly all metrics, with a notable increase of 2.9\% in precision and 1.1\% in F1 score for fake news detection. Compared NSLM, Despite it relying on image tampering and cross-modal consistency for fake news detection, DAAD achieves a 4.7\% improvement by covering a wider range of deceit patterns and using adaptive routing that uses complementary and diverse discriminators.

On the Weibo-21 and GossipCop, despite BMR leverages multi-view features of the news, DAAD achieves better performance through its more comprehensive and adaptive discriminators. On GossipCop, DAAD improves accuracy by 4.8\% over AKA-Fake. While AKA-Fake introduces dynamic knowledge via reinforcement learning, DAAD brings a greater enhancement by integrating dynamic comment from LLMs through prompt optimization. These improvements demonstrate the effectiveness of our adaptive discriminators and the integration of dynamic comments.

\begin{table}[]
\centering
\caption{\small{The ablations of DAAD on Weibo and GossipCop.}}
\vspace{-0.8em}
\resizebox{1.0\columnwidth}{!}{%
\begin{tabular}{c|ccc|ccc}
\toprule
\multirow{3}{*}{Model} & \multicolumn{3}{c|}{Weibo}                             & \multicolumn{3}{c}{GossipCop}                         \\ \cline{2-7} 
                                & \multirow{2}{*}{Acc} & \multicolumn{2}{c|}{F1}   & \multirow{2}{*}{Acc} & \multicolumn{2}{c}{F1}   \\ \cline{3-4} \cline{6-7} 
                                &                               & Fake          & Real          &                               & Fake           & Real        \\ \hline
DAAD                  & \textbf{0.932}                & \textbf{0.928}          & \textbf{0.934}          & \textbf{0.902}                & \textbf{0.734}          & \textbf{0.940}          \\ \hline
DAAD w/o MCTS                  & 0.927                &0.924          & 0.930          & 0.896               & 0.706          & 0.936          \\ \hline
Full                 & 0.924                         & 0.927                   & 0.920                   & 0.894                         & 0.709                   & 0.934                   \\ \hline
w/o RD               & 0.914                         & 0.910                   & 0.916                   & 0.890                         & 0.694                   & 0.932                   \\
w/o FDD                & 0.906                         & 0.909                   & 0.901                   & 0.886                         & 0.684                   & 0.930                   \\
w/o LD                 & 0.907                         & 0.908                   & 0.906                   & 0.888                         & 0.695                   & 0.931                   \\
w/o SD                 & 0.890                         & 0.891                   & 0.890                   & 0.883                         & 0.681                   & 0.928                   \\ \bottomrule
\end{tabular}%
}

\label{tab2}
\vspace{-1.5em}
\end{table}

\subsection{Ablation Study}
We perform a comprehensive ablation study on the Weibo and GossipCop datasets to evaluate the effectiveness of the proposed modules. Additional experiments can be found in Appendix \ref{exper}.

\textbf{Human prompt v.s. Optimized prompt.} Table \ref{tab2} shows the effectiveness of Prompt Optimization. DAAD employs MCTS to optimize prompts, while DAAD w/o MCTS uses manually crafted prompts. Compared to DAAD, DAAD w/o MCTS shows a 0.6\% decrease in accuracy across both datasets and a 2.8\% decrease in F1 score on GossipCop. This demonstrates the importance of crafting domain-specific prompts when integrating comments from LLMs. The limited improvement may be attributed to the constraints of the open-source models used.

\textbf{Discriminator.} 
The impact of discriminators is also shown in Table \ref{tab2}, where `Full' represents the use of all discriminators but not comment from LLM. `w/o RD,' `w/o FDD,' `w/o LD,' and `w/o SD' represent the removal of the ReLU-, Frequency Domain-, Logical-, and Semantic Discriminator, respectively. Compared to `Full,' the performance of `w/o SD' significantly decreases, with  accuracy dropping by 3.4\% and 1.1\% on the two datasets, respectively. This underscores the importance of integrating multimodal features for fake news detection. Additionally, the performance of `w/o FDD' and `w/o LD' also declines on both datasets, indicating that examining the frequency domain and logical consistency of news is effective. Overall, the removal of any discriminator results in a certain degree of performance drop, validating the effectiveness of all four discriminators.

\begin{table}[t]
\caption{\small{The ablations on Weibo for various routing types.}}
\vspace{-0.8em}
\centering
\resizebox{0.36\textwidth}{!}{\begin{tabular}{lccc}
\toprule
\multicolumn{1}{c}{\multirow{2}{*}{Model}} & \multirow{2}{*}{Accuracy} & \multicolumn{2}{c}{F1 Score}    \\ \cline{3-4} 
\multicolumn{1}{c}{}                       &                           & Fake News      & Real News      \\ \hline
Random                                     & 0.902                     & 0.897          & 0.907          \\
Hard                                       & 0.922                     & 0.919          & 0.925          \\
Soft                                       & \textbf{0.932}            & \textbf{0.928} & \textbf{0.934} \\ \bottomrule
\end{tabular}}
\label{tab3}
\vspace{-0.6em}
\end{table}

\begin{table}[t]
\caption{\small{The ablations on the Weibo and GossipCop datasets for different components of Prompt Optimization.}}
\centering
\vspace{-0.8em}
\resizebox{0.42\textwidth}{!}{\begin{tabular}{lcc}
\toprule
\multicolumn{1}{c}{\multirow{2}{*}{Type}} & Weibo          & GossipCop      \\ \cline{2-3} 
\multicolumn{1}{c}{}                      & Accuracy       & Accuracy       \\ \hline
Human                                     & 0.646          & 0.719          \\
MCTS                                      & 0.662          & 0.672          \\
MCTS+Memory                           & 0.692          & 0.759          \\
MCTS+Memory+Resample              & \textbf{0.704} & \textbf{0.777} \\ \bottomrule
\end{tabular}}
\label{tab4}
\vspace{-1.2em}
\end{table}

\textbf{Soft Router.} 
Table \ref{tab3} illustrates the impact of different routing types on the Weibo dataset. `Random' represents random routing, where $\tau$ is sampled from a uniform distribution. `Hard' refers to discrete routing values of $\tau$, meaning the output of the previous discriminator is routed to only one subsequent discriminator. `Soft' indicates adaptive probability values. It can be observed that `Random' and `Hard' show a decrease of 3\% and 2\%, respectively, compared to `Soft', validating the importance of the complementary of the four discriminators.

\textbf{Prompt Optimization.}
Table \ref{tab4} demonstrates the effectiveness of each component in the Prompt Optimization. It is important to note that the accuracy shown here reflects the performance of only using the LLM to classify news. `Human' refers to LLM classification using manually designed prompts. `MCTS' indicates the use of MCTS-optimized prompts for news classification by LLM. `MCTS+Memory' incorporates the MemoryBank component on top of `MCTS'. `MCTS+Memory+Resample' further integrates the Resampling mechanism. The results reveal that using MCTS alone leads to a slight improvement or decrease in accuracy, likely due to the introduction of instance-specific details during optimization, which may reduce generalization. In contrast, `MCTS+Memory' achieves accuracy improvements of 3\% and 8.7\% on the two datasets compared to `MCTS', highlighting the importance of the MemoryBank. The introduction of the Resampling mechanism further enhances performance by 1.2\% and 1.8\% on the two datasets, demonstrating the effectiveness of further exploration around optimized prompts. The prompts obtained with each component are provided in Appendix ~\ref{pro_v}.

\begin{figure}[H]
\vspace{-1.2em}
\centering
\includegraphics[width=0.98\columnwidth]{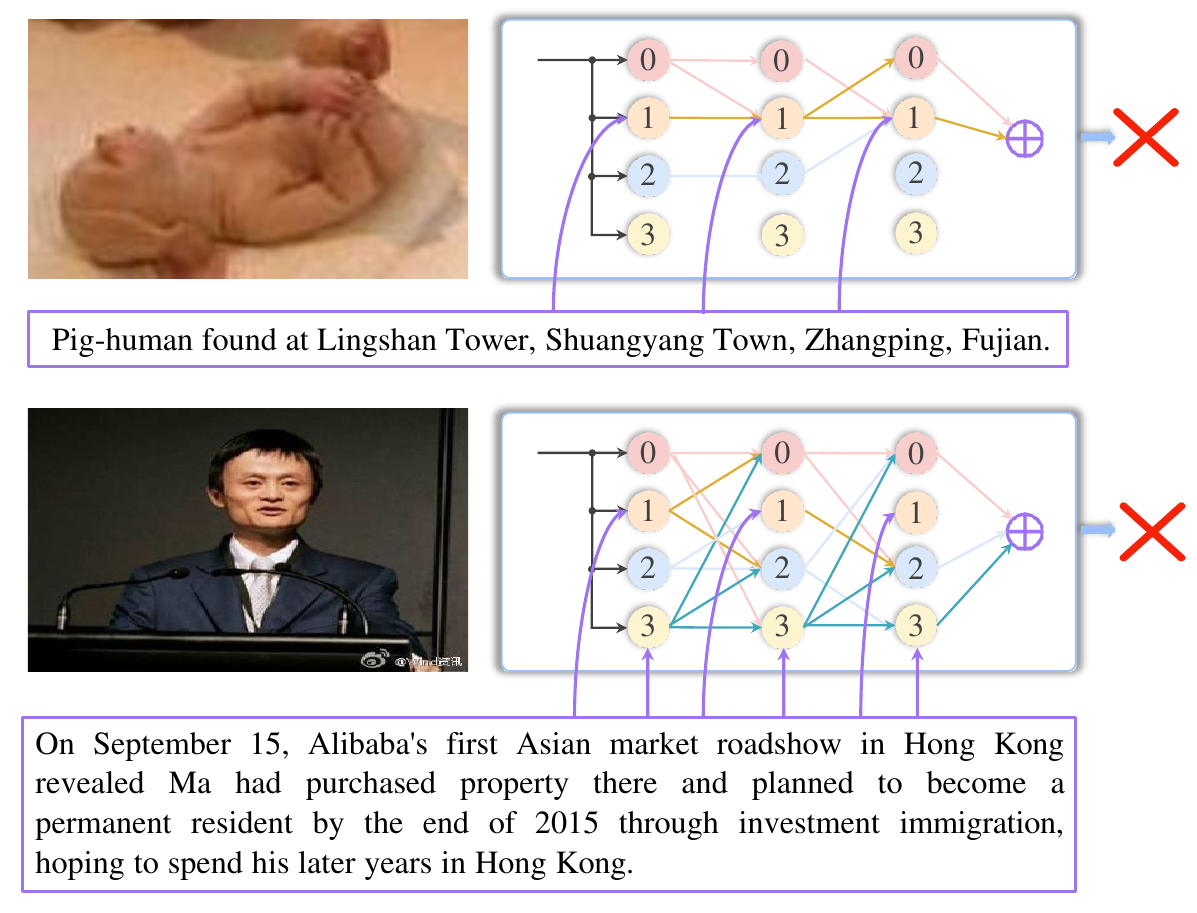} 
\vspace{-1.1em}
\caption{\small{Visualization of routing paths in some fake news examples. The texts are translated from Chinese to English.}}
\label{figvis}
\vspace{-1.0em}
\end{figure}

\textbf{Qualitative results.}
To clearly illustrate the role of each discriminator, we discretized the learned routing using a  $\tau = 0.9$ threshold and visualized the paths for selected fake news cases, as shown in Fig. \ref{figvis}. For blatantly fabricated images (as shown in the top row), the path primarily activates only the ReLU- and frequency domain discriminators. In contrast, for more complex news, the model explores relationships between different discriminators through more intricate paths.

\section{Conclusion}
In this work, we address the issue of insufficient flexibility in existing knowledge-based and semantics-based fake news detection methods by proposing a Dynamic Analysis and Adaptive Discriminator (DAAD) approach. First, we introduce domain-specific comments from Large language models (LLMs) using Monte Carlo Tree Search (MCTS), and mitigated the risk of getting trapped in local minima during optimization through MemoryBank, Batchprompt, and Resampling. Second, we define four typical deceit patterns and design corresponding discriminators, allowing for flexible exploration of optimal detection models through dynamic routing. Finally, extensive experiments on three mainstream datasets demonstrate the superiority of our method.

\section{Limitations}
Although DAAD has demonstrated promising performance, it still has several limitations. First, while domain-specific prompts are introduced through prompt optimization, the meta-prompts used during the optimization process are manually defined, which may lead to suboptimal prompts. Future research could also focus on incorporating more detailed comments tailored to different news domains. Additionally, although various deception patterns and adaptive discriminators have been designed, these are still predefined. Future work could explore how to automatically discover more effective deception patterns and discriminators from the different domain and extend these methods to areas beyond fake news detection, such as Sarcasm and Harmful Meme Detection, aiming to develop a unified detection model.

\section{Ethical Statement}
This paper adheres to the ACM Code of Ethics and Professional Conduct. Firstly, we ensure that the datasets used do not contain sensitive personal information and pose no harm to society. Secondly, any pre-trained models employed have been duly licensed. Furthermore, our code will be released under a suitable license. Lastly, the proposed fake news detection method aims to contribute to the safety and stability of the internet environment and public opinion.
\bibliography{custom}

\clearpage
\appendix

\begin{algorithm*}[h]
\centering
\caption{$\operatorname{Prompt Optimization}(s_0, p_{\mathcal{M}'}, p_{\mathcal{M}''},M, r, d, d_1, L, e,\tau, c)$}
\label{alg:mcts}
\small
\begin{algorithmic}
    \State \textbf{Inputs:}
    \Statex \hspace*{\algorithmicindent} Initial prompt (state) $s_0$, state transition function $p_{\mathcal{M}''}$, reward function $r$, action generation function $p_{\mathcal{M}'}$, depth limit $L$, training epoch $e$, number of generated actions $d$, number of generated state $d_1$ in resampling, MemoryBank $M$, MemoryBank update signal $\tau$, exploration weight $c$ (Equation~\ref{eqn2})

    \State \textbf{Initialize:}
    \Statex \hspace*{\algorithmicindent} State to action mapping $ \mathcal A : \mathcal S \mapsto \mathcal A$, children mapping  or state transition $\text{ch} : \mathcal S \times \mathcal A \mapsto \mathcal S$, rewards $r : \mathcal S \times \mathcal A \mapsto \mathcal R$, State-action value function $Q : \mathcal S \times \mathcal A \mapsto \mathcal R$, visit counter $\mathcal{N} : \mathcal S \mapsto \mathbb N$

    \For {$n \gets 0, \dots, e - 1$}
        \For {$t \gets 0, \dots, L - 1$}
            \If{$s_t$ is not leaf node} \Comment{selection}
                \State $a_t \gets \arg\max_{a \in A(s_t)} \left( Q(s_t, a) + c \cdot \sqrt{\frac{\ln \mathcal{N}(s_t)}{\mathcal{N}(\text{ch}(s_t, a))}} \right)$ 
                \State $s_{t + 1} \gets \text{ch}(s_t, a_t)$, $r_t \gets r(s_t, a_t)$, $\mathcal{N}(s_t) \gets \mathcal{N}(s_t) + 1$ 
            \Else \Comment{expansion and simulation}
                \For {$i \gets 1, \dots, d$}
                \State Sample $a_t^i \sim p_{\mathcal{M}'}(a \mid s_t, \mathcal{M'})$ \Comment{Use Batchprompt}
                \State \text{Update $M$} : $M \gets a_t^i $ \Comment{Use MemoryBank}
                \If{$\tau$ is not false} 
                \State \text{Summarize $M$} 
                \EndIf
                \State $s_{t+1}^i \sim p_{\mathcal{M}''}(s \mid s_t, a_t^i,M, \mathcal{M''})$, and $r_t^i \gets r_\theta(s_t, a_t^i)$
                \State Update $A(s_t) \gets \{a_t^i\}_{i=1}^d$, $\text{ch}(s_t, a_t^i) \gets s_{t+1}^i$, and $r(s_t, a_t^i) \gets r_t^i$
                \EndFor
                \State $a_t \gets \arg\max_{a^i_t \in A(s_t)} r_t^i (s_t, a^i_t)$ 
                \State $s_{t + 1} \gets \text{ch}(s_t, a_t)$, $r_t \gets r(s_t, a_t)$, $\mathcal{N}(s_t) \gets \mathcal{N}(s_t) + 1$ 
                \State $s_{t + 1}^{d_1} \gets \text{Resampling}( s_{t + 1} )$
                \Comment{Use Resampling}
                \State   $s_{t + 1}^{*} \gets \arg\max_{ s \in s_{t + 1}^{d_1} / s_{t + 1} } r_t (s,a_t)$ 
                \State  $s_{t + 1} \gets s_{t + 1}^{*} $ $s_{t + 1} \gets \text{ch}(s_t, a_t)$, $r_t \gets r(s_t, a_t)$ 
                
            \EndIf
            \State \algorithmicif\ {$s_{t+1}$ is an early-stopping state}\ \algorithmicthen\ \textbf{break}
        \EndFor
        \State $T \gets$ the actual number of steps
        \For {$t \gets T - 1, \dots, 0$} \Comment{back-propagation}
            \State Update $Q(s_t, a_t)$ with $\{r_t, r_{t+1}, \dots, r_L\}$ based on Equation~\ref{eq3}
        \EndFor
    \EndFor
\end{algorithmic}
\end{algorithm*}

\section{More Analysis}\label{text-based}
\subsection{Discriminator}

\textbf{ReLU Discriminator:}
For simple fake news detection, complex discriminators are often unnecessary. ReLU functions by mapping negative values to zero while retaining positive values, making it effective in preserving key features. As a result, ReLU can filter out irrelevant information and highlight critical cues like emotional words and exaggerated punctuation. These cues may be sufficient to identify fake news without the need for additional discriminators. Moreover, ReLU helps prevent the vanishing gradient problem and has low computational complexity, making it an efficient choice.

\textbf{Frequency Domain Discriminator:}
The Fast Fourier Transform (FFT) generates a sparse spectrum where many frequency components are close to zero, thereby emphasizing relevant features and suppressing irrelevant ones. Since fake news often involves fabricated facts or manipulated images, spectral analysis can assist in detecting such falsifications. Given that unimodal analysis struggles to detect text forgery or locate image tampering (which is often high-frequency), we utilize text spectra (containing global features) as cues (Eq.~\ref{9}) to identify manipulated or inconsistent information within images. Similarly, visual spectra can help identify fabricated elements in text, such as when text objects are misrepresented relative to image objects.

\textbf{Logical Discriminator:}
Multi-head self-attention captures contextual relationships within local segments (e.g., words or visual regions) across various subspaces, allowing for the detection of grammatical or contextual discrepancies.

\textbf{Semantic Discriminator:}
Cross-attention evaluates semantic consistency between images and text, and Eq.~\ref{eq14} is used to further refine these differences.

In contrast to studies such as NSLM\cite{dong2024unveiling} and BMR\cite{ying2023bootstrapping}, which do not select discriminators based on data type, our approach assigns a router to each discriminator. This allows for the dynamic activation of different pathways based on the input data, enabling multiple discriminators to collaborate effectively and ensure comprehensive detection of various types of fake news. To the best of our knowledge, we are the first to introduce a dynamic mechanism in fake news detection that covers most deception patterns. This dynamic routing mechanism not only optimizes the model’s performance but also allows for the adjustment of the number of discriminators as needed to balance performance and efficiency.

\subsection{Text-based Versions}
In the paper, we implemented an image-based version where the inputs to the ReLU and logic discriminator primarily derive from images. Additionally, we explored a symmetric version focused on text. We defined functions such as \( \mathcal{H}_0(\hat{T}) = \operatorname{ReLU}(\hat{T}) \), \( \mathcal{H}_1(\hat{T}, V ) \), \( \mathcal{H}_2(\hat{T})=\operatorname{FFN}(\text{MultiHead}(\hat{T})) \), and \( \mathcal{H}_3(\hat{T}, V) \), which allow the model to concentrate on the key features of the text. A performance comparison between the two versions is presented in Table \ref{tabvv}. It is important to note that no comments on large language models were introduced in this context.


\begin{table}[]
\centering
\caption{\small{The model variants of DAAD on Weibo and GossipCop.}}
\vspace{-0.8em}
\resizebox{1.0\columnwidth}{!}{%
\begin{tabular}{c|ccc|ccc}
\toprule
\multirow{3}{*}{Model} & \multicolumn{3}{c|}{Weibo}                             & \multicolumn{3}{c}{GossipCop}                         \\ \cline{2-7} 
                                & \multirow{2}{*}{Acc} & \multicolumn{2}{c|}{F1}   & \multirow{2}{*}{Acc} & \multicolumn{2}{c}{F1}   \\ \cline{3-4} \cline{6-7} 
                                &                               & Fake          & Real          &                               & Fake           & Real        \\ \hline
Text-based                 & \textbf{0.929}                & \textbf{0.927}          & \textbf{0.932}          & \textbf{0.896}                &0.706          & \textbf{0.936}          \\ 
Image-basd                 & 0.924                         & \textbf{0.927}                   & 0.920                   & 0.894                         &  \textbf{0.709}                   & 0.934                   \\               \bottomrule
\end{tabular}%
}

\label{tabvv}
\vspace{-1.3em}
\end{table}

\section{Implementation}\label{imple}
We implemented DAAD using PyTorch 2.3.1 and conducted all experiments on a single NVIDIA Tesla A100 GPU. For text and comment feature extraction, we used `bert-base-chinese' for the Weibo and Weibo-21 datasets with a maximum sequence length of 160, and `bert-base-uncased' for the GossipCop dataset with a maximum sequence length of 394. Images were resized to $224 \times 224$ to match the input dimensions of the pre-trained VGG-19 model. In Prompt Optimization, the models $ \mathcal{M} $ for the Chinese and English datasets were implemented with Chinese Llama3 and Llama3.1, respectively, while $\mathcal{M'}$ and $\mathcal{M}''$ were implemented with GPT-3.5. The maximum tree depth was set to 10 and the exploration constant $k$ was set to 2.5, with a batch size of 32, and training was conducted for 16 epochs. In Adaptive Discriminator Modeling, the dimensions of image and text features $d$ were set to 256, with the number of heads set to 4 and a dropout rate of 0.5. We trained the model using AdaBelief~\cite{zhuang2020adabelief} for 50 epochs with a batch size of 32 and an initial learning rate of 1e-4. In addition, for the GossipCop dataset, the hyperparameter $\lambda$ in the adjusted cross-entropy loss (Eq.~\ref{cross}) is set to 3, as discussed in appendix ~\ref{sen}. Additional implementation details can be found in the code.

\section{Addtional Details for Prompt Optimization}\label{po}
\subsection{Implementation Details}
\subsubsection*{Model Versions} 
\begin{itemize}
    \item GPT-3.5: gpt-3.5-turbo-0125
    \item LLama 3: Llama3-8B-Chinese-Chat.  
    \item LLama 3.1: meta-llama/Meta-Llama-3.1-8B-Instruct. 
\end{itemize}
For the Chinese datasets such as  Weibo and Weibo-21, we utilize the Llama3-8B-Chinese-Chat model, which has been fully fine-tuned on both Chinese and English datasets, as the default base LLM. For the English dataset like GossipCop, we use the Llama3.1-8B-Instruct model as the base LLM. To generate actions and states, we require a LLM with strong self-reflection capabilities, therefore, we employ GPT-3.5 for this process. The temperature for the base LLM is set to 0.01, while the temperature for the LLM used in action and state generation is set to 1.0.

\subsubsection*{Prompt Optimization Algorithms}
To clearly illustrate the joint prompt optimization framework based on the Monte Carlo Tree Search (MCTS) algorithm and the MemoryBank, we detail the execution process in Algorithm \ref{alg:mcts}.
\subsection{Meta-prompt}
In this section, we detail the complete format of the specific meta-prompts used in the Markov Decision Process (MDP), MemoryBank, BatchPrompt, and Resampling. Please note that the semantics of the meta-prompts remain unchanged across both the Chinese and English datasets. Only the English meta-prompts are provided here; the Chinese meta-prompts can be obtained by translating the English ones into Chinese.
\subsubsection*{Meta-prompts in MDP}
The `input format' is the actual input for model $\mathcal{M}$ used to generate error examples. The `error string' refers to the format of each error example. The `error feedback' serves as the actual input for model $\mathcal{M'}$, guiding the model to gather error feedback (generate actions). The `state transit' guides the optimizer model $\mathcal{M''}$ to perform state transitions (generate new prompts), which include the sequence of prompts in the selected path (trajectory prompts) and the global guidance (actions) stored in the MemoryBank.

\textbf{initial prompt:}
Please use your knowledge of journalism to determine the authenticity of the input news or social media message, and briefly explain the reason for your judgment.

\textbf{input format:}

\texttt{\{prompt\}} 

News or social media message: \texttt{\{news\}}.

Options: 

A:True 

B. False 

At the end show the answer option between \texttt{<answer>} and \texttt{</answer>}.

\textbf{error string:}

The index of error news or social media message is: \texttt{\textless index\textgreater}

 The model's input is: \texttt{\{input format\}}

 The model's response is:  \texttt{\{response\}}

 The correct label is: \texttt{\{label\}}

 The model's prediction is: \texttt{\{prediction\}}

 \textbf{error feedback:}

 I am writing prompts for a large language model designed to detect fake news.

 My current prompt is: \texttt{\{prompt\}} 

But this prompt gets the following examples wrong: \texttt{\{error strings\}} 

Please carefully examine each wrong example and its incorrect answer. Provide detailed reasons explaining why the prompt leads to the incorrect answer.

Finally, based on these reasons, summarize and list specific  suggestions for improving the prompt according to the following guidelines:

1.The suggestions should focus on how to modify the prompt to help the model classify better, without relying on external knowledge or inputting more information.

2.The suggestions should avoid overfitting specific incorrect examples, ensuring generalizability to different news examples.

\textbf{state transit:}

 I am writing prompts for a large language model designed to detect fake news.

 My current prompt is: \texttt{\{prompt\}} 

But this prompt gets the following examples wrong: \texttt{\{error strings\}}

Based on these errors, the problems with this prompt and the reasons are: \texttt{\{error feedback\}} 

There is a list of former prompts including the current prompt, and each
prompt is modified from its former prompts: \texttt{\{trajectory prompts\}} 

Please consider the following potential causes of model classification errors and suggestions: \texttt{\{Memory\}} 

Based on the above information, please write  \texttt{\{steps\_per\_gradient\}} new prompts following these guidelines:

1. The new prompts should solve the current prompt's problems.

2. The new prompts should consider the list of prompts and evolve based on the current prompt.

3. Each new prompt should be wrapped with \texttt{\textless START\textgreater} and \texttt{\textless END\textgreater}.

The new prompts are:
\subsubsection*{Meta-prompts in MemoryBank}

Summarize the following reasons and suggestions concisely, focusing on key information. If there are multiple key points, summarize them in bullet points. The reason for the error and recommendation is:

Number \texttt{\textless index\textgreater}:

Reasons and Suggestions: \texttt{\{Historical errors\}}

...

\subsubsection*{Meta-prompts in Batchprompt}

To accelerate the process and improve the efficiency of LLM utilization, Batchprompt is used for generating comments and generating error examples.








\textbf{Batchprompt for Comments Generation}
\texttt{\{prompt\}} 

You will receive \texttt{\{batch\_size\}} sentences from the news and social media message dataset as input. The format of the provided sentences is as follows:

Text 1: xxxxxx

Text 2: xxxxxx

...

Please generate the output according to the following format, with the result and reason enclosed in []:

For Text 1: [Result: X. Reason: xxxxxx]

For Text 2: [Result: X. Reason: xxxxxx]

...

Where X is `True' or `False,' and `xxxxxx' is your reasoning for determining the authenticity of the news or social media message.

Ensure that you generate  \texttt{\{batch\_size\}} evaluations. Make sure to output  \texttt{\{batch\_size\}} lines of text, with each line containing a separate evaluation. Do not combine the evaluation results for each text. Ensure that each sentence receives an individual evaluation.

news and social media message \texttt{\textless index\textgreater}:  \texttt{\{news\}}

...

\textbf{Batchprompt for Prompt Optimization}

\texttt{\{prompt\}} 

You will receive \texttt{\{batch\_size\}} questions as input. The format of the provided questions is as follows:

Question 1: xxxxxx

Question 2: xxxxxx

...

Please generate the output according to the following format, with the answer enclosed in []:

For Question 1: [xxxxxx]

For Question 2: [xxxxxx]

...

Ensure that you generate \texttt{\{batch\_size\}} answers, with each question receiving a separate response. Do not combine the answers for each question. Ensure that only one answer is provided for each question.

question \texttt{\textless index\textgreater}:  \texttt{\{input format\}}

...

\subsubsection*{Meta-prompts in Resampling}

Generate \texttt{\{steps\_per\_gradient\}} variants of the prompt \texttt{\{prompt\}} while maintaining the same meaning. Each new prompt should be wrapped with \texttt{\textless START\textgreater} and \texttt{\textless END\textgreater}.

\section{Experiment.}\label{exper}
\subsection{Visualization Experiment}\label{pro_v}
\subsubsection*{Prompt Visualization}
In this section, we present the optimized prompts for the Weibo and GossipCop datasets and provide a visual comparison between the optimized prompts and the standard manually crafted prompt, as shown in Table~\ref{tabc} and Table~\ref{tabe}. Compared to the manually created prompt (highlighted in gray), our optimized prompts offer richer, domain-specific guidance (highlighted in green), and achieve superior performance. However, relying solely on the MCTS algorithm can introduce instance-specific details (highlighted in yellow), which may reduce generalization. This observation is consistent with our analysis in Table~\ref{tab4}.

\subsubsection*{Comment Visualization}
In Table~\ref{tab8}, we visualize the comments generated using optimized prompts. The results show that the comments produced by the LLM are not only effective and aligned with the fake news deceit patterns but also match the guidance provided in our optimized prompts.

Additionally, we visualized the comments generated using both manually crafted prompt and optimized prompt to compare their differences, as shown in Table~\ref{tab9}. The results demonstrate that our domain-specific optimized prompts not only achieve superior performance but also provide reasonable and accurate comments on the news.

\subsection{Parameter Sensitivity Analysis}\label{sen}
Table \ref{tab5} presents the impact of different routing layer numbers on Weibo dataset. The results indicate that moderately increasing the number of $l$ can fully leverage the complementary strengths of the discriminators. However, further increasing the number of routing layers leads to a slight decline in performance, likely due to the increased complexity of the path space, which makes model learning more challenging or causes data overfitting, thereby limiting performance improvements.

\begin{table}[t]
\caption{\small{The results for various routing layers \( l \) on Weibo.}}
\vspace{-0.8em}
\centering
\resizebox{0.31\textwidth}{!}{\begin{tabular}{ccccccc}
\toprule
\multicolumn{1}{c}{\multirow{2}{*}{$\boldsymbol{l}$}} & \multirow{2}{*}{Accuracy} & \multicolumn{2}{c}{F1 Score}    \\ \cline{3-4} 
\multicolumn{1}{c}{}                                  &                                    & Fake News & Real News \\ \hline
1                                            & 0.916                              & 0.914              & 0.919              \\
2                                            & 0.929                              & 0.927              & 0.932              \\
3                                            & \textbf{0.932}                     & \textbf{0.928}     & \textbf{0.934}     \\
4                                            & 0.923                              & 0.920              & 0.925              \\ \bottomrule
\end{tabular}}
\label{tab5}
\vspace{-0.9em}
\end{table}

Table \ref{tab6} presents the impact of different $\lambda$ settings on the loss function in the imbalanced dataset GossipCop. The experimental results indicate that selecting an appropriate $\lambda$ can enhance the model's performance on imbalanced datasets.
\begin{table}[t]
\caption{\small{The results of different $\lambda$ settings on GossipCop.}}
\vspace{-0.8em}
\centering
\resizebox{0.31\textwidth}{!}{\begin{tabular}{ccccccc}
\toprule
\multicolumn{1}{c}{\multirow{2}{*}{$\lambda$}} & \multirow{2}{*}{Accuracy} & \multicolumn{2}{c}{F1 Score}    \\ \cline{3-4} 
\multicolumn{1}{c}{}                                  &                                    & Fake News & Real News \\ \hline
2                                            & 0.903                              & 0.713             & \textbf{0.942}              \\
3                                            & \textbf{0.904}                     & \textbf{0.734}     & 0.940     \\
4                                            & 0.894                              & 0.709              & 0.934              \\ \bottomrule
\end{tabular}}
\label{tab6}
\vspace{-1.9em}
\end{table}
\subsection{Computational Complexity}
The time required for prompt optimization depends on GPT’s response latency, the batch prompt size, the number of resampling iterations per prompt, the number of input tokens, and the sample size per run. Based on the parameters and experimental results presented in the paper, we conducted multiple trials and calculated the average time, which is approximately 15 minutes.

Achieving the performance level reported in our paper requires approximately 4-5 hours of computation on a single Tesla A100 GPU. Table~\ref{tab:params} presents a comparison of the parameters between our method and mainstream methods, demonstrating that our approach attains superior performance while utilizing the fewest parameters.
\begin{table}[h]
\caption{Comparison of model parameters.}
\centering
\vspace{-0.8em}
\resizebox{\linewidth}{!}{ 
\begin{tabular}{lcccc}
\hline
Model & DAAD (Ours) & EANN & SpotFake & MCAN \\
\hline
Params & 124.15M & 143.70M & 124.37M & 248.74M \\
\hline
\end{tabular}
}
\label{tab:params}
\end{table}

\definecolor{lightgreen}{RGB}{204,255,204} 
\definecolor{lightyellow}{RGB}{255,255,204} 
\definecolor{lightgray}{RGB}{211,211,211} 

\begin{table*}
\centering
\caption{Visualization of the prompt optimization process on the Weibo dataset: domain-specific guidance highlighted in green and instance-specific details highlighted in red and bold. The text is translated from Chinese to English.}
\vspace{-0.8em}
\begin{tabular}{p{3.8cm}p{11cm}p{0.6cm}} 
\toprule
Approach &  Prompt & Acc. \\ \hline
Human & \sethlcolor{lightgray}\hl{Please use your knowledge of the news to determine the authenticity of the input news or social media message, and provide a brief explanation of the reasoning behind your judgment.} & 0.646 \\ 
MTCS & Based on the specific content of the given news or social media message,\sethlcolor{lightgreen}\hl{ assess its authenticity by considering keywords, the context of the event, and relevant evidence, and explain your reasoning.} When determining truthfulness, \sethlcolor{lightyellow}\hl{pay attention to whether the message \textbf{involves personal privacy, police actions, or content released by authoritative organizations. For sensational or unusual events}, go beyond subjective impressions by thoroughly analyzing and verifying the information sources.} \sethlcolor{lightgreen}\hl{Ensure that you avoid subjective speculation and base your judgment on objective facts and evidence.} In your response, provide detailed explanations of your reasoning to improve accuracy. & 0.662 \\ 
MCTS+Memory & Please assess the authenticity and reasonableness of the message content, \sethlcolor{lightgreen}\hl{including whether it aligns with common sense and logic, whether it is supported by detailed information, whether it comes from a credible source, and whether there might be a motive to spread false information.} Based on these factors, determine the truthfulness of the message and briefly explain your reasoning. & 0.692 \\ 
MCTS+Memory+Resample & Based on the content of the provided news or social media message, \sethlcolor{lightgreen}\hl{and by considering relevant background knowledge,} assess its authenticity and provide a detailed explanation of your reasoning.  \sethlcolor{lightgreen}\hl{In making your judgment, pay close attention to the logical consistency of the message content, the reliability of the information source, the authority of the message's publisher, and any potential signs of fraud or characteristics of false information.} & 0.704 \\ \bottomrule
\end{tabular}
\label{tabc}
\end{table*}

\begin{table*}
\centering
\caption{Visualization of the prompt optimization process on the GossipCop dataset: domain-specific guidance highlighted in green and instance-specific details highlighted in red and bold.}
\vspace{-0.8em}
\begin{tabular}{p{3.8cm}p{11cm}p{0.6cm}} 
\toprule
Approach &  Prompt & Acc. \\ \hline
Human & \sethlcolor{lightgray}\hl{Please use your knowledge of the news to determine the authenticity of the input news or social media message, and provide a brief explanation of the reasoning behind your judgment.} & 0.719 \\ 
MTCS & \sethlcolor{lightgreen}\hl{Please analyze the given news or social media message for emotional language, logical content, and credibility.} \sethlcolor{lightyellow}\hl{Consider the context of the news, \textbf{such as promotional strategies in the entertainment industry or common trends in reality TV shows}, to determine the overall truthfulness of the information presented.} \sethlcolor{lightgreen}\hl{Evaluate the publisher's motive, the reasonableness of the content, and the presence of detailed information like time, place, and actions to make an informed judgment.} & 0.672 \\ 
MCTS+Memory & Please assess the authenticity and reasonableness of the message content, \sethlcolor{lightgreen}\hl{including whether it aligns with common sense and logic, whether it is supported by detailed information, whether it comes from a credible source, and whether there might be a motive to spread false information.} Based on these factors, determine the truthfulness of the message and briefly explain your reasoning. & 0.759 \\ 
MCTS+Memory+Resample & \sethlcolor{lightgreen}\hl{Please analyze the authenticity of the news or social media message by focusing on the reported events, details, and factual accuracy. Consider the relevance of emotional language, logical consistency, and credibility in determining the truthfulness of the content. Pay attention to the motive of the publisher, the reasonableness of the information provided, and the presence of specific details like time, location, and actions to make an informed judgment on the reliability of the message.} & 0.777 \\ \bottomrule
\end{tabular}
\label{tabe}
\end{table*}

\definecolor{lightred}{RGB}{255, 182, 193}  
\definecolor{lightblue}{RGB}{173, 216, 230} 
\definecolor{lightpurple}{RGB}{216, 191, 216} 

\begin{table*}
\centering
\caption{Visualization of comments generated by the large language model using optimized prompts on the Weibo dataset. The text is translated from Chinese to English. Different colors represent different analytical perspectives. Exaggeration (highlighted in red), Lack of or inclusion of specific details (highlighted in blue), Consistency or inconsistency with common sense (highlighted in purple).}
\vspace{-0.8em}
\begin{tabular}{p{8.8cm}p{6.6cm}} 
\toprule
News & Comment \\ \hline
Recently, a 2.3 billion yuan high-speed rail project in Jilin was exposed for using stones instead of concrete in its bridge piers, sparking alarm! International media has exclaimed that "riding China's high-speed rail is like committing suicide"! Recently, netizens in a southern province have also discovered that the high-speed rail bridge piers exposed by dried-up riverbeds were similarly built with stones instead of concrete! A billion-dollar high-speed rail project hides a huge disaster black hole! & \sethlcolor{lightgray}\hl{Low credibility} . \sethlcolor{lightred}\hl{The news title and content are exaggerated and emotional, using strong terms such as "black hole" and "suicide",}  \sethlcolor{lightblue}\hl{lacking specific accident data and official confirmation.} This suggests a possible attempt to mislead the public for clickbait.  \\ 
Due to 2 dollars, special police beat up a driver, who then knocked them out (KO) with one move! Seems like they died. Don’t be too arrogant outside, you never know who the BOSS is! & \sethlcolor{lightgray}\hl{Unreliable}. \sethlcolor{lightred}\hl{The headline is exaggerated, using dramatic terms like “beating up” and “KO,”} \sethlcolor{lightblue}\hl{and lacks specific details and factual evidence.} It is likely a piece of online rumor or false information.  \\ 
$[$Jiangsu mandates a complete stop to outpatient IV infusions in all secondary and higher-level hospitals! Are you still overusing antibiotics?$]$ Recently, Jiangsu Province introduced a new regulation requiring all secondary and higher-level hospitals to completely stop outpatient intravenous infusions by the end of 2016, except for children's hospitals. When you come across these antibiotics, use them reasonably and cautiously! Spread the word! (People's Daily) & \sethlcolor{lightgray}\hl{Result: True.} \sethlcolor{lightblue}\hl{Reason: The information comes from People's Daily, a highly credible official newspaper in China. }\sethlcolor{lightpurple}\hl{Additionally, the mandate for hospitals to stop outpatient IV infusions is a health policy that aligns with public health principles and logic. }\\ 
Latest News from CNN: Malaysian Airlines flight shot down by the Malaysian Air Force over the Malacca Strait. [Link to webpage] & \sethlcolor{lightgray}\hl{Result: True}. \sethlcolor{lightblue}\hl{Reason: The source of the news is CNN, a well-known news organization known for its rigorous news verification processes. The report involves an international event, specifying the time, location, and entities involved, which increases its credibility. Additionally, a webpage link is provided as evidence, which is typical of news reports.} Therefore, based on the credibility of the source and the evidence provided, the news is likely to be true.  \\ \bottomrule
\end{tabular}
\label{tab8}
\end{table*}

\begin{table*}
\centering
\caption{Comparison of performance in news classification using original and optimized prompts on the Weibo dataset: labels (highlighted in gray), classification errors (highlighted in red), and correct classifications (highlighted in green). The optimized prompt not only achieves superior performance but also provides insightful commentary on the news. The text is translated from Chinese to English.}
\vspace{-0.8em}
\renewcommand{\arraystretch}{0.56} 
\resizebox{\textwidth}{!}{
\begin{tabular}{p{6.2cm}p{0.5cm}p{3.8cm}p{5.7cm}} 
\toprule
News &  Label & Original Prompt & Optimized Prompt \\ \hline
[Extra Discounts on All Brands: Year-End Sale on Top Brands] Still showing off your Double Eleven purchases? Put down your phone if you're running out of money—the real discounts are here! @Time Outlets 6th Anniversary Celebration, from November 27th to 28th, enjoy an additional 5\% off on already discounted items starting from 30\% off the original price; from November 27th to December 6th, the entire store will have discounts starting at 30\% off. Get incredible deals on top brands like Coach, Burberry, and more—stock up while you can! & \sethlcolor{lightgray}\hl{True} & \sethlcolor{lightred}\hl{Unreliable.} The news uses exaggerated language and exclamation marks, lacks specific details and sources, and may be a commercial advertisement or false information. & \sethlcolor{lightgreen}\hl{Result: True.} Reason: This message appears to be a commercial advertisement promoting upcoming discounts at a shopping venue. It provides specific dates, discount details, and mentions well-known brands, which are typical features of advertisements. The message does not mention any unreasonable or unlikely situations, so it can be considered a genuine promotional message. \\ 
$[$DIY Non-Toxic Mosquito Repellent: A Must-Have for Parents$]$ Many commercial mosquito repellents and floral waters contain DEET, a chemical often used as a pesticide, which is not suitable for infants and young children. Here’s a recommended method to make your own non-toxic mosquito repellent: Take a small spray bottle, add five tablets of vitamin B complex, fill the bottle with mineral water, and shake well. Spray it on the body, as mosquitoes are said to be repelled by the scent of vitamin B complex. & \sethlcolor{lightgray}\hl{False} & \sethlcolor{lightred}\hl{Result: True.} Reason: Commercial mosquito repellents may contain ingredients unsuitable for infants, making a homemade non-toxic repellent a safer option for parents. & \sethlcolor{lightgreen}\hl{Credibility is low.} Reason: The news mentions DEET, which is not commonly found in many commercial mosquito repellents, and lacks specific data or sources to support this claim. Additionally, the effectiveness of vitamin B complex in repelling mosquitoes lacks scientific evidence. \\ 
$[$Duang! Search Assistant Is Back with a Giveaway! 20 Movie Tickets Up for Grabs$]$ Recent blockbusters like The Martian, Dragon Blade, and Mr. Six are plentiful! Are you ready? Search Assistant is giving away tickets again. Follow @SichuanFoodSearch and retweet to three friends for a chance to win 2 Taobao movie vouchers! Note: Download the @Taobao Movies app and use the vouchers for online seat selection. See you tonight at 8 PM! & \sethlcolor{lightgray}\hl{True} & \sethlcolor{lightred}\hl{Result: False.} Reason: This text appears to be a promotional advertisement designed to attract users to follow and retweet in order to receive benefits, mentioning Search Assistant, movie tickets, and the Taobao Movies app. & \sethlcolor{lightgreen}\hl{Result: True.} Reason: The message is consistent with common sense and logic, as it describes a plausible activity where a social media account (@SichuanFoodSearch) is running a promotion with movie tickets and Taobao movie vouchers as prizes. It provides specific participation steps and prize details and advises users to download the Taobao Movies app, indicating that the activity is likely real and feasible. Although the source is not an authoritative organization, as a social media account with a specific follower base, the message's authenticity is relatively high.

 \\ 
$[$Russian Man Develops Tree in Lung After Inhaling Seed$]$ Seeds can root and sprout in the soil, but have you ever heard of a plant growing inside a human body? Russian doctors recently discovered a 5 cm (2-inch) pine tree growing in a man's lung, which had developed from a pine seed he inhaled earlier. The man reported severe pain but did not feel any foreign object inside his body. & \sethlcolor{lightgray}\hl{False} & 
\sethlcolor{lightred}\hl{Result: True.} Reason: This text describes a peculiar medical situation, providing specific details and background information, and includes a credible source marker "Global Network" , which increases the likelihood of its authenticity. & \sethlcolor{lightgreen}\hl{Result: False.} Reason: The credibility is low. Firstly, the possibility of a tree growing inside the human body is extremely unlikely and biologically implausible. Secondly, the claim that the patient did not feel the foreign object might be due to the lack of severe health issues from the seed. Such reports are likely exaggerated to attract attention. \\ \bottomrule

\end{tabular}
}
\label{tab9}
\end{table*}


\algnewcommand{\Inputs}[1]{%
  \State \textbf{Inputs:}
  \Statex \hspace*{\algorithmicindent}\parbox[t]{0.98\linewidth}{\raggedright #1}
}
\algnewcommand{\Initialize}[1]{%
  \State \textbf{Initialize:}
  \Statex \hspace*{\algorithmicindent}\parbox[t]{0.98\linewidth}{\raggedright #1}
}

\end{document}